\documentclass[conference]{IEEEtran}

\newif\ifcameraready
\camerareadyfalse

\usepackage{cite}
\usepackage{amsmath,amssymb,amsfonts}
\usepackage{algorithmic}
\usepackage{graphicx}
\usepackage{textcomp}
\usepackage{xcolor}
\usepackage{booktabs}
\usepackage{tipa}
\usepackage{hyperref}
\def\BibTeX{{\rm B\kern-.05em{\sc i\kern-.025em b}\kern-.08em
    T\kern-.1667em\lower.7ex\hbox{E}\kern-.125emX}}
\begin{document}

\title{Benchmarking Human and Automatic Speech Recognition of Diverse Speech: Initial Results}

\author{
    \IEEEauthorblockN{
        Ilse Huisman\textsuperscript{*},
        Rares Popa\textsuperscript{*},
        Yuanyuan Zhang, 
        Odette Scharenborg
    }
    \IEEEauthorblockA{
        \textit{Multimedia Computing Group} \\
        \textit{Delft University of Technology}\\
        Delft, the Netherlands \\
        \{y.zhang-44, o.e.scharenborg\}@tudelft.nl\\
    }
}

\maketitle
\begingroup
\renewcommand{\thefootnote}{*}
\footnotetext[1]{These authors contributed equally.}
\endgroup

\begin{abstract}
Humans are often considered to be the best listeners and seen as the upper-bound performance of automatic speech recognition (ASR) systems. We present a preliminary comparison of the performances of state-of-the-art ASR systems and Dutch native listeners on the recognition of ``diverse'' speech, specifically Dutch child and older adults' speech and Flemish. Google Telephony outperformed the other ASR systems. Importantly, the ASR systems showed similar performance to the listeners, and in specific cases even outperformed them. Slight performance differences between the listeners and ASR systems were found related to speaker’s age and regional accents and utterance length. Future research should focus on making ASR systems more robust to acoustic variability related to aging and regional accents. A comparison of ASR recognition performances on the test stimuli and the full Jasmin-CGN test sets showed the influence of the specific test sets on the conclusions regarding benchmarking human and ASR performance.
\end{abstract}

\section{Introduction}

Automatic speech recognition (ASR) systems have come a long way from being a magnitude worse in their performance than human listeners as Lippmann showed in his seminal work comparing human listeners and ASR systems from 1997~\cite{Lippmann1997}. He concluded that ASR performance could be improved if research would focus on improving acoustic-phonetic modelling, increasing robustness against noise and channel variability, and on modelling spontaneous speech. 
ASR systems have since then regularly been benchmarked against human listeners to investigate performance differences in the recognition of phonemes~\cite{Cooke2008}, logotomes~\cite{Meyer2011, Meyer2013}, and words in quiet and background noise ~\cite{Cooke2008, Meyer2011, Meyer2013,  Spille2018, Juneja2012} in order to identify areas of potential improvement of current ASR systems.

In 2017, ASR performance was shown to be on par with human transcription performance on two conversational English databases~\cite{Xiong_humanparity2017}; however, human performance was obtained from professional transcribers who have been found to have a much better performance than quick transcription, which is closer to everyday listening conditions~\cite{glenn-etal-2010-transcription}. More recently, in 2024, Patman and colleagues~\cite{Patman2024} showed that human listeners consistently outperformed Wav2vec 2.0~\cite{Baevski2020} for speech produced by speakers with and without wearing a face mask in noise, while Whisper-large~\cite{Whisper_large_v3} was found to outperform the human listeners in all conditions except pub noise. Comparisons of the recognition errors of ASR systems and human listeners have highlighted both similarities and differences~\cite{Xiong_humanparity2017,Lopez2022,Mansfield2021}.

Human listeners are often seen as the upper-bound performance of ASR systems. However, the above results show that current state-of-the-art (SotA) ASR systems not only reach very good performance for different listening conditions, they in fact in specific conditions outperform human listeners on standard speech. At the same time, increasingly more research has shown that ASR performance is significantly worse for speakers who deviate from the ``standard'' native, adult speaker of a language without a strong accent and speech impediment. For instance, child speech is recognised worse than adult speech (e.g., ~\cite{Qian2017,Feng2024,Herygers2022, Narayanan2022}) and atypical speech, e.g., due to dysarthria ~\cite{Sanders2002, xiong2019phonetic, Liu2021, hernandez22_interspeech, alsayegh2025zero, derussis2019}, oral cancer~\cite{Halpern2022} or cleft lip and palate~\cite{Schuster2006}, is worse recognised than typical speech.  Moreover, non-native ~\cite{Wu2020,Palanica2019,Zhang2022} and regionally-accented  ~\cite{Feng2024,Herygers2022,Koenecke2020,tatman2017ACL,Tatman2017Interspeech,arabic_accents2013} speech are recognised worse than ``standard'' speech. 

This paper presents preliminary results investigating the question how SotA ASR systems compare to human listeners on the recognition of non-standard or ``diverse'' conversational speech. We compare recognition performance of native Dutch listeners to that of three off-the-shelf SotA ASR systems on Dutch speech from three diverse speaker groups: child speech, speech from older adults, and Dutch as spoken in Flanders (i.e., Flemish). We aim to identify areas for potential improvement of current ASR system's ability to deal with the acoustic variability of diverse speech to make them more inclusive, i.e., accessible to everyone irrespective of how one speaks or the language one speaks.
\section{Methodology}

We ran three separate listening experiments (see Section~\ref{subsec:experiments}) to investigate the recognition performance of human listeners on child speech, older adults' speech, and Flemish teenager and older adults' speech and compared their recognition results with the recognition performances (see Section~\ref{subsec:eval}) of Google Telephony~\cite{Google_speech_models}, Whisper-large-v3~\cite{Whisper_large_v3}, and a custom Dutch Conformer model (see Section~\ref{subsec:ASR}), on the exact same stimuli (see Section~\ref{subsec:stimuli}).

\subsection{Stimuli}
\label{subsec:stimuli}

We selected 120 stimuli from the child, older adults and Flemish speakers from the Jasmin-CGN corpus~\cite{cucchiarini2006}, which is a corpus of spoken Dutch from people living in the Netherlands and the Flanders region. For each speaker group, we selected 40 stimuli from the human-computer interaction (HMI) test set of the corpus, as this type of speech most closely resembles conversational speech. For each experiment, we balanced the speech samples as much as possible over different demographic labels. In all three experiments, 20 utterances were spoken by female speakers and 20 by male speakers. Additionally:

\textbf{Child speech (7-11 years).} For each age, four stimuli were selected of different utterance length, with 1 utterance of 5-6 words, 1 of 7-8 words, 1 of 9-11 words, and 1 utterance of 12-16 words (excluding non-speech sounds; mean word count: 8.0, SD: 2.4). Regional accents were balanced as much as possible.

\textbf{Older adults' (OA) speech (59-96 years).} To capture the wide variability of regional accents, 5 male and 5 female speakers were selected from each of the 4 dialect regions in Jasmin-CGN. Age was distributed as good as possible over the regions and genders. Each utterance consisted of at least 4 words (excluding non-speech; mean word count: 6.3, SD: 1.8). 

\textbf{Flemish speech (13-17 and 65-84 years).} We tested both teenagers and older adults. Ten stimuli were selected from each of the 4 dialect regions, within each region balanced for age group and gender. Each utterance consisted of 6-16 words (excluding non-speech; mean word count: 7.3, SD: 1.2).

\subsection{Human listening experiments}
\label{subsec:experiments}
\subsubsection{Participants}
Forty-five adult native listeners of Dutch were recruited from the social and work circles of the experimenters. Twenty listeners (10 female and 10 male participants, average age: 39.5, SD: 9.5) participated in the child speech experiment\footnote{Originally the participants were split into two groups: ten listeners with and 10 listeners without experience with child speech, however, no performance difference was found between the two groups, hence, we report the aggregated results here.}; 14 participants in the experiment with older adult speech (7 female and 7 male, average age: 37.2, SD: 16.7), and 11 in the experiment with Flemish speech (5 female and 6 male, average age 40.1, SD: 13.6). No participants reported hearing problems. Participants did not receive payment for their participation. Each listener participated only in one experiment, except for two participants who participated in both the child speech and older adult speech experiments. The study was approved by the ethical committee of our university.

\subsubsection{Experimental set-up}
All stimuli were normalized to the same volume using the loudnorm filter of FFmpeg~\cite{FFmpegLoudnorm}. 
The experiments were hosted in Qualtrics, ran on the same laptop, and used the same headphones (Sennheiser HD 200 Pro) within an experiment. All experiments were held in a quiet room. Prior to the experiments, listeners received instructions, signed a consent form, and had a short familiarisation with the experiment of one stimulus which was also used to adjust the volume to a comfortable level. During the experiment, one stimulus was shown per page. The order of the stimuli was randomized per participant to prevent order effects. Listeners were allowed to listen to each stimulus only once to mimic everyday listening conditions; however, in the Flemish experiment, listeners were allowed to listen twice due to a different setting of the experiment.

Participants were asked to type what they heard. After every 10 stimuli, listeners were allowed to take a break for as long as they liked. Correct transcriptions were not shown during the experiment because this could induce learning due to the pop-out effect~\cite{Davis2005}, which could influence the results. Participants were allowed to compare their answers to the correct transcriptions after conclusion of the experiment, which many did. 

\subsection{Automatic speech recognition systems}
\label{subsec:ASR}
The three models used in this study were selected from the best performing systems in a recent study which compared 12 SotA ASR systems for Dutch on the diverse speech of Jasmin-CGN ~\cite{zhang2026}. The first model is Google Telephony, which, in line with the HMI speech used in this study, is optimised for conversational, telephone speech. Google Telephony was the second best model (after Google Chirp). We performed synchronous speech recognition using the Google Cloud API via the 2.33.0 version of the Speech-to-text V2 python package. 
Whisper-large-v3 (Whisper) is an often used baseline model. It was the 4th-to-6th best model (depending on the speech type) in~\cite{zhang2026}. Whisper was downloaded from HuggingFace. During testing, we employed beam search decoding with a beam size of 10. We set the task to ``transcribe'', the language to ``Dutch'', and the temperature parameter to 0.
The third model is a custom Conformer model using XLSR-53 features~\cite{Conneau2021XLSR} trained with ${\sim}700$ hours of Dutch standard speech from the Corpus Gesproken Nederlands (CGN; ~\cite{oostdijk-2000-spoken}), which obtained similar results to Whisper in~\cite{zhang2026}.

\subsection{Evaluation}
\label{subsec:eval}
To make the human transcriptions in line with those of the ASR systems, all transcriptions were converted to lowercase, punctuation was removed (except for the apostrophe), any additional spaces were removed, digits/numbers were written out in full, and all non-linguistic symbols, filler words, and other transcriptions of non-lexical sounds were removed from the transcriptions of both the human listeners and the ASR systems. Obvious typing errors and spelling mistakes were corrected when the intended word was fully unambiguous (e.g., ``hius'' $\rightarrow$ ``huis'' (house); ``afentoe'' $\rightarrow$ ``af en toe'' (sometimes)).

Recognition performances of the human listeners and the ASR systems were measured in Word Error Rate (WER). For the comparison of the performance of the human listeners and the three ASR systems, a paired bootstrap was used with 10000 speaker-based resamples and 95\% confidence intervals (CI)~\cite{Bisani2004, ferrer2024good}. A difference was considered statistically significant if CI excludes zero. P-values are only reported when the results are significant. Furthermore, we compared the type of errors made by the human listeners and the ASR systems. Since~\cite{Mansfield2021} found that human errors are highly correlated with the speaker, and large performance disparities have been observed by ASR systems for different speaker groups, we also investigated the effect of the speakers' age, reported gender, and regional accent on the human listeners' and ASR results. For statistical significance testing, a multi-factor linear model $$\text{WER}\_{\text{spk}} \sim \text{model} \times (\text{gender} + \text{regional}\_\text{accents})$$  using the estimated marginal means package (emmeans)~\cite{lenth2023emmeans} was employed in R to explore the effect of gender and regional accents. Moreover, we investigated the effect of the number of words in an utterance on the WERs. 
\section{Results}

\subsection{Recognition performance for the diverse speech}

Table \ref{tab:diverse_results} shows the WERs of the human listeners and the three ASR models for the three experiments on child speech, older adults' speech, and Flemish (split for teen(ager) and older adults' speech and averaged over the full stimuli set). For the human listeners, the standard deviations (SDs) per speaker group are provided. These show that while for child speech the human listeners were fairly consistent in their responses (and errors), for the other speaker groups, the SD is quite a bit higher, indicating larger variability in the WERs for the individual participants. Regarding the ASR systems, no significant differences were found between the three models. 

Comparing the performances of the human listeners and the ASR models, for child speech, there are no significant differences in the performance of the human listeners and the ASR systems. For older adults' speech, there was a significant difference between the human listeners and the ASR systems: both the Google Telephony model (95\% CI [+3.426\%, +16.148\%], \textit{p} = .0031) and Whisper (95\% CI [+2.454\%, +12.610\%], \textit{p} = .0041) 
 outperformed the human listeners. For Flemish, on average, all three ASR systems outperformed the human listeners. Splitting the results into teenager (20 stimuli) and older adults' speech (20 stimuli) for easier comparison with the Dutch child and older adults' results, for teenager speech, the Google Telephony model significantly outperformed not only Whisper (95\% CI [+2.381\%, +17.901\%], \textit{p} = 0.0067) but also the human listeners (95\% CI [+6.935\%, +13.228\%], $\textit{p}<0.001$). For Flemish older adults' speech, there was no significant difference between the human listeners and the ASR systems.

\begin{table}[ht]
    \centering
    \caption{WER (\%) of the human listeners and three ASR models on the 40 stimuli of child speech, older adults' (OA) speech, and Flemish. For the human listeners, the standard deviation of the WER is also given. Bold indicates the best results for a speaker group.}
    \label{tab:diverse_results}
    \resizebox{\linewidth}{!}{
    \begin{tabular}{c|c|c|c|c|c}
        \toprule
         &  \multicolumn{2}{c|}{\textbf{Dutch}} & \multicolumn{3}{c}{\textbf{Flemish}}  \\ 
         & \textit{Child} & \textit{OA} & \textit{Teen} & \textit{OA} & \textit{Avg} \\ \hline
        Human listeners & 17.6$\pm$3.0 & 26.8$\pm$7.5 & 20.9$\pm$8.3 & 17.7$\pm$7.4  & 19.3$\pm$7.6  \\ \hline
        \midrule
        Google Telephony & \textbf{12.8}  & \textbf{17.1} & \textbf{10.3} & \textbf{10.4} & \textbf{10.3}  \\ \hline
        Whisper & 16.0 & 19.1 & 20.5 & 13.2 & 16.9 \\ \hline
        Conformer & 18.5 & 22.0 & 15.1 & 11.8 & 13.4 \\
        \bottomrule
    \end{tabular} \label{tab:WERs}
    }
\end{table}

\subsection{Analysis of the error types}
\label{subsec:errortypes}
Analysis of the insertions, deletions, and substitution patterns showed that for all three speaker groups, both the humans and the ASR systems showed very few insertions. In all cases, the majority of the errors were substitutions, accounting for 8.9\%-13.2\% of the WER for child speech, 11.3-14.2\% for older adults' speech, and 7.1-13.0\% of the WER for Flemish. The Conformer model had the highest substitution rate for child and older adults' speech, while Whisper had the highest rate for Flemish, closely followed by the human listeners (11.3\%). For child speech and older adults' speech, the human listeners showed a relatively high deletion rate (5.3\% and 9.5\%). This suggests that human listeners do not always write something down when they do not understand what has been said, while ASR systems will recognise something when there is audio. These results are somewhat counter those of ~\cite{Lopez2022} who observed far more deletions for the ASR systems, but these concerned primarily discourse markers, while our stimuli were selected to not contain these. Note however that~\cite{Mansfield2021} found the opposite pattern, with human listeners being likelier to miss discourse markers. 

Analysis of the most common substitutions showed that both the human listeners and ASR systems made errors involving \textit{me} (English: \textit{me}), \textit{mijn}, and its reduced form \textit{'m} (English: \textit{mine}), which are not only acoustically very similar, but also semantically highly related. Similar substitutions were found for \textit{ik} and its reduced form \textit{'k} (English: \textit{I}) and \textit{het} and its reduced form \textit{'t} (English: \textit{it}). Normalising these transcriptions, reduced the WERs for the human listeners and ASR systems by between 6-8\% but did not change the patterns presented in Table~\ref{tab:WERs}.


\subsection{The effect of demographic variability}

\subsubsection{The effect of the age of the speaker}
\begin{figure}
    \centering
    \includegraphics[width=1.0\linewidth]{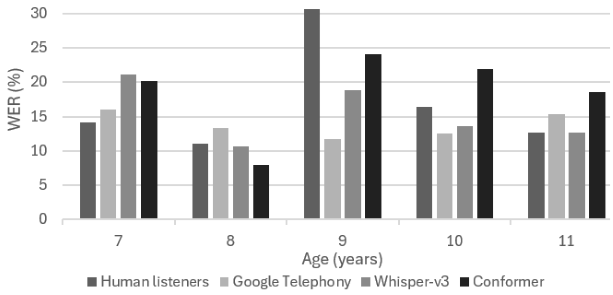}
    \caption{WER (\%) of the human listeners and ASR systems split for the age of the child speakers.}
    \label{fig:age_children}
\end{figure}

Figure \ref{fig:age_children} shows the WER of the human listeners and ASR systems split for the child speakers' ages. Since each bin only contains 10 stimuli, no statistical tests were carried out and no hard conclusions can be drawn; nevertheless, a trend can be observed, which seems to suggest that recognition performance does not improve nor deteriorate with age for the human listeners and the ASR systems.  Further analysis of the high WER for the 9-year old speakers, particularly for the human listeners, showed that this is primarily due to a single harder-to-recognise speaker.

\begin{figure}
    \centering
    \includegraphics[width=1.0\linewidth]{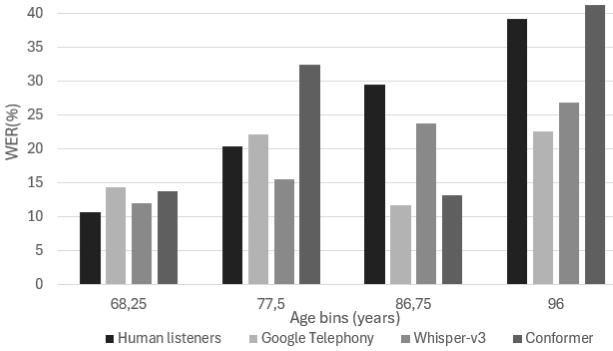}
    \caption{WER (\%) of the human listeners and ASR systems split per age bin for the older adult speakers.}
    \label{fig:age_OA}
\end{figure}

For the older adults' speech, Figure \ref{fig:age_OA} suggests an age trend, with increasing WER for the human listeners and the ASR systems with increasing age (age is binned into 4 bins), and particularly so for the human listeners and the Conformer model. Google Telephony and Whisper seem to be slightly more robust against acoustic variability due to aging. However, also here the number of stimuli per bin is low. Future research will need to investigate these trends further.

\subsubsection{The effect of the gender of the speaker}
Table \ref{tab:gender-region} shows the WER differences between male and female speakers for the human listeners and the ASR systems for the child, older adults, and Flemish speakers. For the child and older adults' speakers, the observed performance disparities for the human listeners and the ASR systems were non-significant. For Flemish, while the human listeners and Google Telephony again did not show a significant gender gap, Whisper (t(40)=-2.149, \textit{p}=0.0378) and the Conformer model (t(40)=-3.131, \textit{p}=0.0033) did.

\if 
[B.2] Gender Comparison WITHIN Each Model (Uncorrected):
model\_name = averaged\_human:
 contrast estimate     SE df t.ratio p.value
 F - M     -0.0201 0.0669 40  -0.300  0.7660

model\_name = conformer:
 contrast estimate     SE df t.ratio p.value
 F - M     -0.2095 0.0669 40  -3.131  0.0033

model\_name = google\_telephony:
 contrast estimate     SE df t.ratio p.value
 F - M     -0.1198 0.0669 40  -1.790  0.0810

model\_name = whisper:
 contrast estimate     SE df t.ratio p.value
 F - M     -0.1438 0.0669 40  -2.149  0.0378
\fi

\begin{table}[ht]
    \centering
    \caption{WER differences (\%) for the gender of the speaker (male-female) and regional accents (highest and lowest WER).}
    \label{tab:gender-region}
    \resizebox{\linewidth}{!}{
    \begin{tabular}{c|c|c|c|c|c|c}
        \toprule
         &  \multicolumn{3}{c|}{\textbf{Gender}} & \multicolumn{3}{c}{\textbf{Regional accents}}  \\ 
         & \textit{Child} & \textit{OA} & \textit{Flemish} & \textit{Child} & \textit{OA} & \textit{Flemish} \\ \hline
        Human listeners  & 5.1  & 0.7   & 3.1   & 5.8   & 14.7   &  21.2     \\ \hline      
        \midrule
        Google Telephony & 4.2  & -3.3  & 13.1  & 7.6   & 15.4   &  16.3       \\ \hline 
        Whisper       & 5.5  & 1.1   & 15.3  & 15.9  & 15.0   &  20.3      \\ \hline     
        Conformer        & 10.2 & -11.7 & 24.0  & 1.8   & 33.8   &  23.6       \\ 
        \bottomrule
    \end{tabular}
    }
\end{table}

\if
model\_name = averaged\_human:
 model term       df1 df2 F.ratio p.value
 regional\_accents   2  88   0.437  0.6471

model\_name = conformer:
 model term       df1 df2 F.ratio p.value
 regional\_accents   2  88   0.393  0.6763

model\_name = google\_telephony:
 model term       df1 df2 F.ratio p.value
 regional\_accents   2  88   1.538  0.2205

model\_name = whisper:
 model term       df1 df2 F.ratio p.value
 regional\_accents   2  88   1.148  0.3221
 \fi
 \if
model\_name = averaged\_human:
 model term       df1 df2 F.ratio p.value
 regional\_accents   3 140   3.260  0.0235

model\_name = conformer:
 model term       df1 df2 F.ratio p.value
 regional\_accents   3 140   2.056  0.1088

model\_name = google\_telephony:
 model term       df1 df2 F.ratio p.value
 regional\_accents   3 140   1.215  0.3068

model\_name = whisper:
 model term       df1 df2 F.ratio p.value
 regional\_accents   3 140   2.542  0.0588
 \fi
 \if
 model\_name = averaged\_human:
 model term       df1 df2 F.ratio p.value
 regional\_accents   3  40   2.394  0.0827

model\_name = conformer:
 model term       df1 df2 F.ratio p.value
 regional\_accents   3  40   1.519  0.2244

model\_name = google\_telephony:
 model term       df1 df2 F.ratio p.value
 regional\_accents   3  40   0.881  0.4591

model\_name = whisper:
 model term       df1 df2 F.ratio p.value
 regional\_accents   3  40   1.261  0.3007
 \fi
 
\subsubsection{The effect of regional accent of the speaker}
Table \ref{tab:gender-region} shows the WER difference between the region with the highest and the lowest WER to investigate the effect of regional accent on recognition performance. For the child speech experiment, speech from three regions was used, while for the older adults and Flemish speech experiments speech from four regions was used. For the child speech, none of the (small) performance difference between the regions with the highest and the lowest WER were significant. Similarly, despite the much larger WER differences between the regions for Flemish speech, none of these differences were significant. Only for the older adults, the performance difference for the human listeners was found to be significant (F(3, 140)=3.260, \textit{p}=0.0235), most likely due to a stronger regional accent for the older adults. A post hoc Tukey test showed that the speakers from the West region were significantly better recognised than those from the South region (t(140)=-2.855, \textit{p} = 0.0252), which is not surprising as most listeners were from the West region. The ASR systems did not show an effect of regional accent.




\subsection{The influence of utterance length}
\begin{figure}
    \centering
    \includegraphics[width=1.0\linewidth]{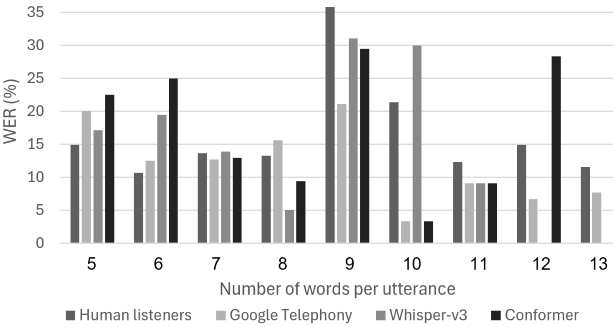}
    \caption{WER per utterance length for the child speech.}
    \label{fig:utt-length-child}
\end{figure}

\begin{figure}
    \centering
    \includegraphics[width=1.0\linewidth]{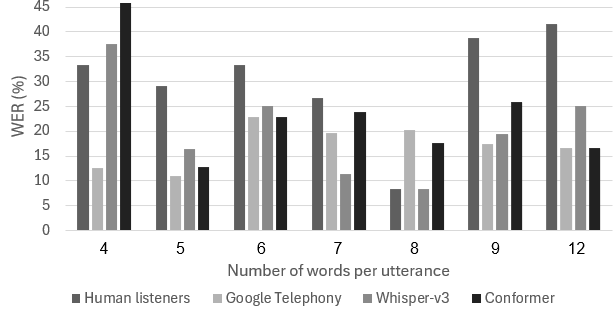}
    \caption{WER per utterance length for the older adults' speech.}
    \label{fig:utt-length-OA}
\end{figure}

Figures \ref{fig:utt-length-child} and \ref{fig:utt-length-OA} show the WERs split by the number of words in the utterance for the child speech and older adults' speech, respectively. For the child speech, human listeners did not seem to show an effect of utterance length on WER; however, for the ASR systems, there is a downward trend for longer utterances. This performance improvement for the ASR systems is potentially due to the longer-range context, which better leverages the language modelling capabilities of large models. This was also observed for Whisper in~\cite{zhang2026}. For the older adult speakers, we observe a downward trend for the Whisper and the Conformer models, but not so for the human listeners and the Google Telephony model. Note however that the number of stimuli with the higher number of words is low (four utterances with length 9 and one with length 12). Overall, there is no (strong) effect of utterance length.

\subsection{ASR results on the full Jasmin-CGN test sets}
To investigate the generalisability of the ASR results beyond the used stimuli, Table \ref{tab:ASR-fulltest} shows the WER for the three models on the full test sets of the child speech (3526 utterances, 1.45 hours), older adults' speech (8292 utterances, 3.77 hours), and Flemish speech (5209 utterances, 2.79 hours) of Jasmin-CGN. Comparison with the results for the 40 stimuli (Table \ref{tab:diverse_results}) shows that the order of the models' performance remains highly similar. Nevertheless, we see that the WERs on the full test sets are on average 8.4\% higher than those for the experimental stimuli. Comparing these results with those of the human listeners in Table~\ref{tab:WERs} on the smaller set of only 40 stimuli shows no significant differences between the performances of  the human listeners and the ASR systems. 

\begin{table}[ht]
    \centering
    \caption{WER (\%) of the ASR models on the full child speech, older adults' speech, and Flemish speech test sets of Jasmin-CGN. Bold indicates the best results for a speaker group.}
    \label{tab:ASR-fulltest}
    \resizebox{0.8\linewidth}{!}{
    \begin{tabular}{c|c|c|c|c}
        \toprule
        \textbf{ASR} &  \multicolumn{2}{c|}{\textbf{Dutch}} & \multicolumn{2}{c}{\textbf{Flemish}}  \\ 
         & \textit{Child} & \textit{OA} & \textit{Teen} & \textit{OA} \\ \hline
        Google Telephony &  \textbf{20.3} &  \textbf{24.1} &  \textbf{19.9} &  \textbf{22.5} \\ \hline
        Whisper & 27.0 & 29.3 & 29.6 & 24.9 \\ \hline
        Conformer & 27.8  & 29.1 & 27.1 & 22.8 \\
        \bottomrule
    \end{tabular}
    }
\end{table}


\section{General discussion and conclusion}
We presented preliminary results addressing the question how state-of-the-art ASR systems compare to human listeners on the recognition of non-standard or ``diverse'' conversational Dutch speech. Specifically, we compared recognition performance of native Dutch listeners with those of Google Telephony and Whisper and a custom-trained Conformer model for Dutch child speech and older adults' speech, and Flemish. Overall Google Telephony obtained the best results of the three ASR systems\footnote{Google Telephony significantly outperformed Whisper and the custom Conformer model for several speaker groups. Since the training data for both Google Telephony and Whisper are unknown, we cannot exclude the possibility that Google Telephony was trained with CGN-Jasmin data. However, the CGN-Jasmin data provider requests disclosure explicitly, which was not made for Google Telephony.} and, surprisingly, significantly outperformed the human listeners for older adults' speech and Flemish teenager speech; moreover, Whisper outperformed the human listeners for older adults' speech. We thus extend the results of~\cite{Xiong_humanparity2017} showing parity of ASR and human recognition performance of standard speech to diverse speech. We show that ASR systems reach parity with human performance for child speech and some systems even exceed human performance for older adults' and Flemish speech. So while human listeners are often seen as the upper-bound performance of ASR systems, our results show that ASR systems in fact in specific conditions not only can outperform human listeners on standard~\cite{Patman2024} speech but also diverse speech. 

Analysis of the effect of demographic variability of the speakers on recognition performance showed a possible effect of age for older speakers, with increasing WER for both the human listeners and the ASR systems with increasing age, in line with~\cite{vipperla_ageing_2010}. No effect of age was found for child speech in our small sample. Since the stimuli and listeners used in the three experiments were different, we cannot draw strong conclusions from a comparison across different speaker groups; nevertheless, it is somewhat surprising to see that generally speaking the speech of the older adults was recognised worst by the ASR systems (except for Flemish older adults' speech) and the human listeners, while child speech, which acoustically deviates quite substantially from adult speech was overall recognised best. These results are opposite those reported in the literature for Dutch for a hybrid TDNNF-HMM model~\cite{Feng2024}, Wav2Vec 2.0~\cite{fuckner2023uncovering} and Whisper-large-v2~\cite{fuckner2023uncovering}, which all had higher WERs for child speech than for older adults' speech. Future research will need to show whether this is due to the chosen stimuli or whether this is a general effect. Nevertheless, these results show that more recent ASR models have improved on the recognition of particularly child speech, making them more inclusive for this speaker group.

The human listeners did not show an effect of gender, which was also the case for the ASR systems, except for Whisper and the Conformer model for the Flemish speech. These results are largely in contrast to previous findings in the ASR literature (e.g.,~\cite{Feng2024,Herygers2022, Koenecke2020,tatman2017ACL,Tatman2017Interspeech,arabic_accents2013,fuckner2023uncovering}), which found gender effects for ASR systems, but in line with recent results from SotA ASR systems and models~\cite{zhang2026}, which only found significant gender-related performance disparities for Whisper. An effect of regional accent was only found for human listeners for older adults' speech, not for the ASR systems, which is in contrast to earlier findings~\cite{Feng2024,Herygers2022}. These results show that recent ASR systems have become more robust against acoustic variation due to regional accents, potentially due to larger variation in the large amounts of training data that are used for training the SotA systems.

A comparison of the WER of the ASR systems on the experimental sets and the full Jasmin-CGN diverse speech tests showed that the test stimuli are an important factor: different or more stimuli lead to different conclusions on the performance gap between human listeners and ASR systems. Future research will need to focus on different stimuli and larger test set sizes to further investigate the performance differences and similarities between human listeners and ASR systems. Moreover, for better comparison of the results for the different speaker groups, a similar number of participants from the general population, and an identical experimental set-up for each speaker group will be used in future research.

In conclusion, while human listeners are often seen as the best listeners, this does not mean they make no recognition errors. In fact, their error rates on the diverse speech in this study are similar to and in some cases significantly higher than those of SotA ASR systems. Nevertheless, overall performance of the ASR systems on diverse speech falls short compared to typical speech. Although ASR systems have improved a lot since Lippmann's seminal paper~\cite{Lippmann1997}, acoustic-phonetic modelling still requires attention. Particular areas for further improvement of ASR systems -- to make them more inclusive -- should focus on improving the modelling of the acoustic variability due to demographic variation related to, specifically, age and regional accents.

\section{Acknowledgements}
We are grateful for the contributions of Ansen Weng to the listening experiments regarding the speech of the older adults.

No generative AI was used for the writing of this paper. 
ChatGPT was used for drafting debugging Python scripts for preprocessing the speech files and selecting the stimuli sets.

\bibliographystyle{IEEEtran}
\bibliography{IEEEabrv,mybib}

\end{document}